\documentclass[12pt]{article}

\usepackage{amsmath, amssymb, amsfonts, amsbsy, amsthm, latexsym, color}
\usepackage{algorithm}
\usepackage{algorithmic}
\usepackage{graphicx}
\usepackage{enumerate}
\everymath{\displaystyle}
\numberwithin{equation}{section}

\begin{document}

\title{A survey on deep supervised hashing methods for image retrieval }

\author{
	Xiaopeng ~Zhang \\
	Department of Mathematics \\
	Nankai University \\
	\texttt{zhangxiaopeng@mail.nankai.edu.cn}\\
	}

\maketitle
\begin{abstract}
Hashing has been widely used in approximate nearest search for large-scale database retrieval for its computation and storage efficiency.Deep hashing, which devises convolutional neural network architecture to exploit and extract the semantic information or feature of images, has received increasing attention recently. In this survey, several deep supervised hashing methods for image retrieval are evaluated and I conclude three main different directions for deep supervised hashing methods. Several comments are made at the end. Moreover, to break through the bottleneck of the existing hashing methods, I propose a $Shadow$ $Recurrent$ $Hashing$(SRH) method as a try. Specifically, I devise a CNN architecture to extract the semantic features of images and design a loss function to encourage similar images projected close. To this end, I propose a concept: "shadow" of the CNN output. During optimization process, the CNN output and its shadow are guiding each other so as to achieve the optimal solution as much as possible. Several experiments on dataset CIFAR-10 show the satisfying performance of SRH.
\end{abstract}

\vspace{0.35in}
\section{Introduction}

With the explosive growing of data in real applications, approximate nearest neighbor (ANN) search has attracted much attention in machine learning, with a lot of applications in information retrieval, computer vision and so on.
Given a query, our goal is to retrieve images that are similar to the query, where ``similar" could refer to visually similar or semantically similar. In fact, it is not that hard if we do not make any restriction. The simplest way of looking for similar images is to find every feature of images (output of the AlexNet fc7, for example) and rank them according to the distances to the query in the feature space. However, with millions of images on the Internet, we can hardly do so for the high storage consuming and expensive searching time. For a database with millions of images, which is quite common in today's world, even a linear scan through the database would cost a great deal of time and memory. To address the inefficiency of the linear searching, recently a widely used technique for ANN search, hashing, has attracted lots of intention. Hashing methods represents images by binary codes instead of real-valued features, thus the time costs from searching and retrieval can be largely reduced.

The image retrieval approaches can be categorized into two main streams, namely, classical retrieval approaches using local features and deep methods based on CNNs producing feature extractor and hash functions. Most existing classical methods basically extract features from hand-crafted extractors, thus the performance of such methods could heavily depends on the feature they use or the approaches to extract feature they design. So for classical methods for fast image retrieval, they are more suitable for dealing with visual similarity rather than semantically similarity. Moreover, most classical hashing methods only perform well on specific problems for the corresponding feature extractors are designed temporarily. While traditional hashing methods have showed their successful performances on some real world datasets, CNNs based methods demonstrate more impressive learning power recently, especially in cases where the local features are not good enough. These deep methods have shown the state-of-the-art performance. In this survey, we mainly focus on deep hashing methods for image retrieval and find out those pros and cons for them. Moreover, inspired by those deep hashing methods, I propose a new method on large scale image retrieval and test the performance on CIFAR-10 dataset.

\section{Extensive evaluation}
While the very first hashing methods based on deep learning should be dated back to semantic hashing Hinton et.al.[1], this method imposes difficult optimization and its performance is surpassed by lots of classical hashing methods. After a decade, CNN began to show much power in several applications and soon a CNN based deep hashing method for image retrieval was proposed.

\begin{figure}[htbp]
\centering
\includegraphics[height=4cm,width=16cm]{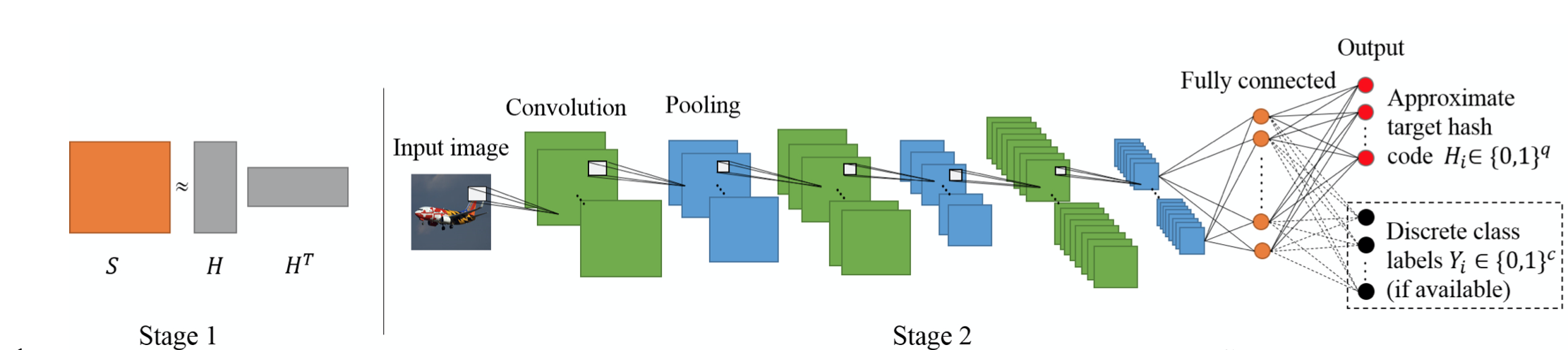}
\caption{model architecture of CNNH}
\end{figure}

\subsection{Convolutional Neural Network Hashing}
Convolutional Neural Network Hashing(CNNH) [2] is the first approach to make full use of the CNN architecture. It first proposes a scalable coordinate descent method to decompose the pairwise similarity matrix $S$ into  $HH^{T}$, where $H$ is a matrix with each of its row being the approximate hash code associated to a training image. Then feature representations of training images are learnt through a CNN architecture as well as a set of hash functions. Moreover, the author further improve the stage of hash functions learning by incorporating the image class labels as a part in the output layer.

Figure1 shows the overview of CNNH. It consists of two main stages. In stage1, approximate hash codes are calculated by minimizing the following reconstruction errors:
$$\min_{H} \sum_{i=1}^{n}\sum_{j=1}^{n}(S_{i,j}-\frac{1}{q}H_{i} \cdot H_{j}^{T} )^{2}=\min_{H} \left \| S-\frac{1}{q}HH^{T}\right \| _{F}^{2}$$

where $H$ is $n\times q$ matrix, with each row being the target q-bit hash code of image $I_{k}$. $S_{i,j}=1$ when $I_{i}, I_{j}$ are semantically similar, $S_{i,j}=-1$ respectively. In stage2, CNNH simultaneously learn a feature extractor for query images as well as a set of hash functions.Herein, CNNH leverages the convolutional neural network to take advantage of the 2D structure of images and to learn the underlying features, and then uses these features to control the outcomes of the corresponding hash codes.

CNNH devises a CNN architecture in which the CNN can be viewed as a feature extractor and capture the underlying semantic structure of images in spite of significant appearance variations. From this aspect, it surpasses most classical hashing methods. However, since it is the first try with the CNN architecture, it inevitably has some drawbacks. In this two-stage method, the learned approximate hash codes are used to guide the learning of the image representation, but the learned image representation cannot give feedback for learning better approximate hash codes. It seems like these two stages are in a bit of reverse order.

One should note that in supervised learning, similarity set $S=\{s_{ij}\}$ can be constructed from semantic labels of data points or relevance feedback from click-through data in real retrieval systems. Although CNNH is not an end-to-end method, the requirement of the similarity matrix should not be viewed as a main drawback. In fact, such matrix is involved as the supervised information in most of the deep supervised methods.

\subsection{Deep neural network hashing}

In CNNH, the approximate hash codes are determined at the very beginning of the whole process, which seems to be contradictory to our common logic. The approximate hash codes guide the learning process of hash function, which fails to give any feedback to the hash codes learning process. In other words, a pair of semantically similar or dissimilar images may not have feature vectors with relatively small or large Euclidean distance. This one-way interaction is limited and needs to be changed.  

\begin{figure}[htbp]
\centering
\includegraphics[height=6cm,width=8cm]{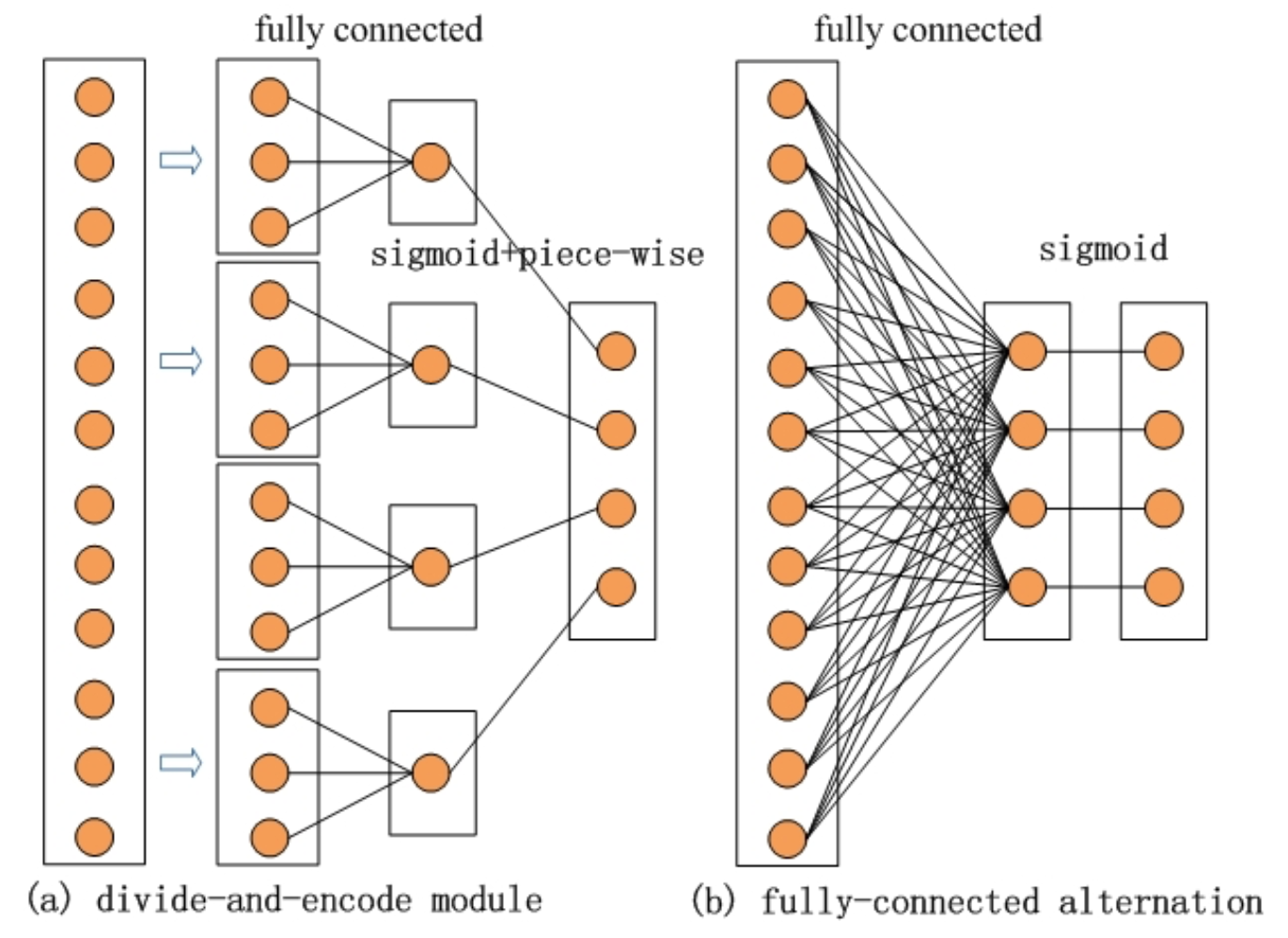}
\caption{(a)A divide-and-encode approach \quad (b)A fully-connected layer and a sigmoid layer}
\end{figure}

Deep Neural Network Hashing(DNNH) [3] generates hash codes of images through a CNN-based architecture. Throughout the deep network, input images are converted into unified image representations via a shared subnetwork of stacked convolution layers. Then, as shown in figure2, these intermediate image representations are encoded into hash codes by divide-and-encode procedure. Specifically, the input intermediate features are divided into $q$ slices, with each of the slices mapped to one dimension by a fully connected layer. Furthermore, a triple ranking loss is designed to preserve relative similarities as well as pulling dissimilar pairs far away.

This work makes a few designs on the network architecture. First, at the end of the full-connected layer, each hash bit is generated from a separated slice of features in order to reduce the redundancy among the hash bits. Second, in order to encourage the output of a divide-and-encode module to be binary codes, a sigmoid activation function is used followed by a piece-wise threshold function. Third, a triplet loss function is designed to characterize that two images are similar and other two are dissimilar. Unlike CNNH, this one-way method plans as a whole for the two learning stage and make sure the two stages are instructing each other so that the coding process is compatible with the feature-extraction process.

\subsection{Deep supervised hashing}
Deep Supervised Hashing(DSH) [4] constructs a CNN architecture and elaborately designs a loss function to maximize the discriminability of the Hamming space by encoding the supervised information from the input image, then imposes regularization on the real-valued outputs to approximate the desired discrete values. As shown in figure3, DSH devises a CNN model which processes pairs of images to determine whether two images are similar and then generates the corresponding hash code with relatively small distance. The loss function is designed to make sure similar query points are projected to close outputs in Hamming space 
\begin{figure}[htbp]
\centering
\includegraphics[height=5cm,width=16cm]{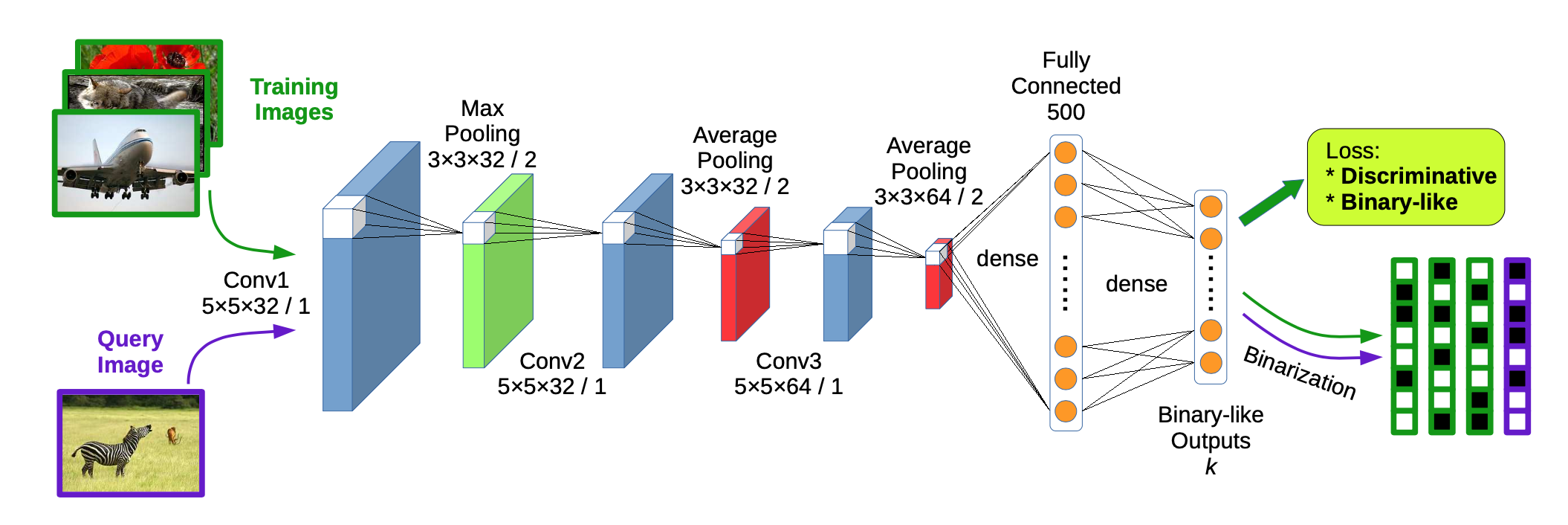}
\caption{model architecture of DSH}
\end{figure}
and push the outputs of the dissimilar ones far away. The loss function is defined as:$$L(b_{1},b_{2})=\frac{1}{2}(1-y)D_{h}(b_{1},b_{2})+\frac{1}{2}y\max(m-D_{h}(b_{1},b_{2}),0)$$   $$s.t. \quad b_{j}\in \{+1,-1\} ^{k},j\in \{1,2\}$$

where $m > 0$ is a margin threshold parameter. The first term punishes similar pairs when projected far away. The second term punishes dissimilar pairs projected close when their distance fall in the threshold $m$. Suppose N pairs are trained, the loss function is defined as :$$L=\sum_{i=1}^{N}L(b_{i,1},b_{i,2},y_{i})$$ $$s.t.\quad b_{i,j}\in \{+1,-1\}^{k},i\in \{1,..,N\},j\in \{1,2\}$$.

Herein, we cannot directly optimize this loss function due to the discrete constraints. A common relaxation method is to use $tanh$ or $sigmoid$ function to approximate the discrete constraints. However, working with such non-linear function would slow down the network. DSH designs the loss function by adding a regularizer as follows.
$$L(b_{1},b_{2})=\frac{1}{2}(1-y)\left \| b_{1}-b_{2}\right \| _{2}^{2}+\frac{1}{2}y\max (m-\left \| b_{1}-b_{2}\right \|_{2}^{2},0)+\alpha (\left \|  \left| b_{1}\right|-1 \right \|_{1}+\left \| \left|b_{2}\right|-1 \right \|_{1})$$

In this work, a novel idea is to impose an additional regularizer to replace the binary constraints and substitute the sigmoid function or tanh function, non-linear functions restraining  and slowing the convergence of the network, especially when the output is near the boundary. The proposed regularizer overcome the problem because its gradient is always 1. More importantly, DSH is an end-to-end architecture. The learned approximate hashing codes are used to guide the learning of the image representation and the learned representation gives feedback to the learning of hash codes.
\subsection{Deep Anchor Graph Hashing}
Deep Anchor Graph Hashing(DAGH) [5] mainly addresses the mini-batch issue. Due to the high computation cost and limited hardware's memory, lots of deep supervised hashing methods will first select a subset from the training set, and then form a mini-batch to update the network in each iteration. Therefore, the remaining labeled data cannot be fully utilized and the model cannot directly obtain the binary codes of the entire training set for retrieval. This method regards the samples in subsets as anchors and design a regression term to establish the connections between the anchors and all the binary codes. A training subset is selected in every iteration and is used to update the neural network by back-propagation.

The DAGH gives a try to tackle the mini-batch issue by using the anchor graph to characterize the similarities between the deep features and all the binary codes. But we should note that this issue is caused by CNNs, whether or not the time-consuming is less by this work which actually do not improve the CNNs architecture should be tested further. However, we should admit that this work provides us with a new perspective towards the mini-batch issue.

\subsection{Deep Incremental Hashing Network} 
Although the mini-batch issue is difficult to solve, Deep Incremental Hashing Network(DIHN) [6] open a new path to somehow tackle the issue. DIHN learns the hash codes for the new coming images directly, while keeping the old ones unchanged. We should remind that the incremental hashing codes are not simply quantified through the original deep hashing network. Instead, DIHN first devises a incremental hashing loss to preserve the semantic similarity between training points, and then learns the new incremental binary codes by integrating the CNN learning model into the incremental hashing loss. DIHN is the first deep incremental hashing approach, It simultaneously generate hash codes for incremental database images and learn a CNN model for producing hash codes for query images. Predictably, DIHN performs well on large scale retrieval system and decrease the training time with no accuracy decline.

\subsection{Asymmetric Deep Supervised Hashing}
Most deep supervised hashing methods adopt a symmetric strategy to learn one deep hash function for both query points and database points. The training of these symmetric deep supervised hashing methods is typically time-consuming. The storage and computational cost for these hashing methods with pairwise supervised information is $\mathcal{O}(n^{2})$ where n is the number of database points. Besides the requirement of mini-batch for CNNs, most deep supervised hashing methods subject to the time-consuming hashing learning procedure so that only a small subset from the whole database are sampled. Asymmetric Deep Supervised Hashing(ADSH) [7] treats the query points and database points in an asymmetric way. ADSH learns a deep hashing function only for query points, while the hash codes for database points are directly learned. The computational cost of ADSH is $\mathcal{O}(n)$.

\begin{figure}[htbp]
\centering
\includegraphics[height=4cm,width=10cm]{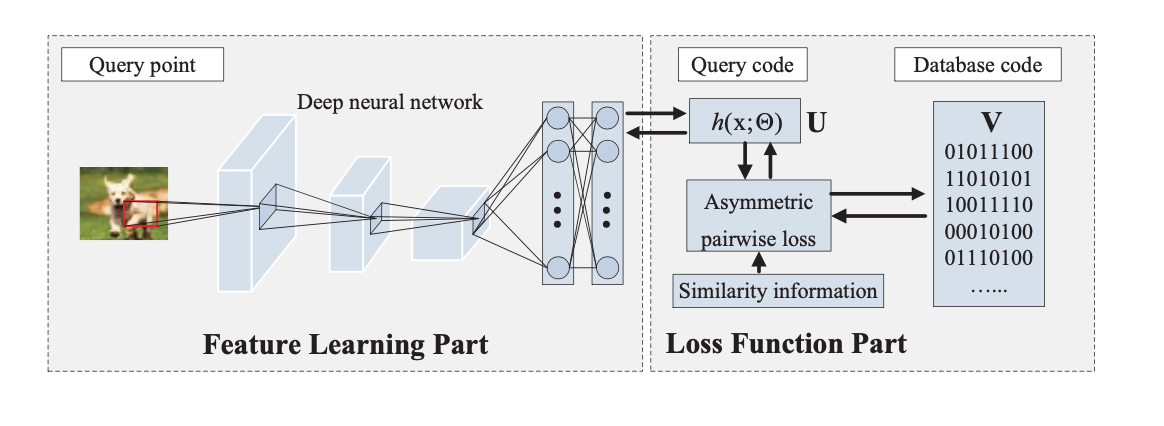}
\caption{model architecture of ADSH}
\end{figure}

As shown in Figure4, the model architecture of ADSH consists of two parts, namely feature learning part and loss function part. Unlike previous hashing methods, ADSH assumes that the whole database contains $m$ query points $X=\{x_{i}\}_{i=1}^{m}$ and $n$ database points $Y=\{u_{j}\}_{j=1}^{n}$. The corresponding binary hash code are denoted as $U=\{u_{i}\}_{i=1}^{m}\in \{-1,+1\}^{m\times k}$ and $V=\{v_{j}\}_{j=1}^{n}\in \{-1,+1\}^{n\times c}$ . In the feature learning part, ADSH adopt the CNN architecture to extract semantic feature from the query images and leverages the outputs to construct the following loss function.$$\min_{U,V}L(U,V)=\sum_{i=1}^{m}\sum_{j=1}^{n}(u_{i}^{T}v_{j}-cS_{i,j})^{2}$$
$$s.t. \quad U\in \{-1,+1\}^{m\times c},V\in \{-1,+1\}^{n\times c},u_{i}=h(x_{i}),i=1,2..,m$$ However, it is difficult to optimize $h(x_{i})$ due to the discrete constraints, so ADSH sets $h(x_{i})=sign(F(x_{i},\mathcal{O}))$, where $\mathcal{O}$ is the parameter of the CNN model. Moreover, ADSH adds a regularizer to keep $v_{i}$ and $tanh(F(y_{i},\mathcal{O}))$ as close as possible. So the final loss function is designed as:$$\min_{\mathcal{O},V}=\sum_{i=1}^{m}\sum_{j=1}^{n}[tanh(F(y_{i};\mathcal{O}))^{T}v_{j}-cS_{i,j}]^{2}+\gamma \sum_{i=1}^{m}[v_{i}-tanh(F(y_{i};\mathcal{O}))]^{2}$$ $$s.t.\quad V\in \{-1,+1\}^{n\times c}$$

In the loss function stage, the learning algorithm consists of two stages. In the first steps, we learn $\mathcal{O}$ with $V$ fixed. By back propagation algorithm, the neural network parameters $\mathcal{O}$ is updated. In the second stage, we learn $V$ with $\mathcal{O}$ fixed. So the problem can be rewritten as $$\min_{V} L(V)=\left \| \tilde{U}V^{T}-cS\right \|^{2}_{F}+\gamma \left \|V -\tilde{U}\right \|^{2}_{F}=\left \| \tilde{U}V^{T}\right \|^{2}_{F}-2ctr(V^{T}S^{T}\tilde{U})-2\gamma tr(V\tilde{U^{T}})$$ $$s.t.\quad V\in \{-1,+1\}^{n\times c}$$ . We can get the optimal solution as: $$V_{*k}=-sign(2\tilde{V_{k}}\tilde{U^{T}_{k}}\tilde{U_{*k}}+Q_{*k})$$ where, $U_{*k}$ denotes the $k$th column of $U$ and $\tilde{U_{k}} $ denotes the matrix of $U$ excluding $U_{*k}$

Although it seems incredible to reduce the computational complexity so much, we have to notice that the feature of the database points are not taken into account throughout the whole algorithm since the hash codes for database points are directly learned without the CNNs extracting the features. As a result, the binary codes for database points are not compact with respect to those query points. However, this work gives us a new perspective. And a possible improvement for this work can be adding an additional procedure in which the database points and the query points reverse their positions. A whole new loss function is probably able to be given. What's more, the computational cost should still be O(n). We can go on looking for this trying then.

\subsection{HashNet}
HashNet [8] mainly focuses on two problems in deep supervised hashing, ill-posed problem and data imbalanced problem.

First, most deep learning hash methods have to give a continuous representation through CNNs and generate binary hash codes in separated step, which would result in suboptimal binary codes. HashNet addresses this problem by gradually reducing the amount of smoothing during the training, which results in a sequence of optimization problems converging to the original optimization problem. Second, the similarity information is usually very sparse in real retrieval systems. The number of similar pairs is much smaller than the number of dissimilar pairs. This will result in the data imbalanced problem, making similarity-preserving ineffective. HashNet tackles the problem by weighting the training pairs according to the importance of misclassifying the pair.

HashNet addresses deep learning to hash from imbalanced similarity data by a continuation method. Admittedly, this continuation method cannot be applied to large-scale dataset since it will definitely cost too much running time. On the other hand, it is better to adopt this method to check out those updates during every iteration in most deep supervised hashing methods.

\subsection{Deep Cauchy Hashing}
Based on HashNet, Deep Cauchy Hashing(DCH) [9] improves the Hamming space retrieval by designing a pairwise cross-entropy loss based on Cauchy distribution, which penalizes significantly on similar image pairs with Hamming distance larger than the given Hamming radius threshold. Most state-of-the-art deep hashing methods usually adopt generalize sigmoid function as the probability function. However, we can observe that the probability of generalized sigmoid function stays high when the Hamming distance between hash codes is much larger than 2 and only starts to decrease obviously when the Hamming distance becomes close to $K/2$. DCH solves the misspecification problem of sigmoid function by constructing the probability function based on the Cauchy distribution instead of the sigmoid function. As we know, for large-scale dataset, to quantize the distances among those compact and concentrated hash codes is significant. The DCH approach enables more efficient and effective Hamming space retrieval based on a Bayesian learning framework. However, the drawbacks of DCH is also apparent. To compensate the efficient retrieval among concentrated hash codes, most of the distances in Hamming spaces are probability stay quite large, thus obscuring the original image retrieval.

DCH mainly focuses on concentrating relevant images to be within a small Hamming ball, thus enables the most efficient constant-time search, which is Hamming space retrieval, instead of using linear scan. DCH first proposed a Bayesian learning framework to perform deep hashing on query images by quantizing the similarity of pairwise images. The goal is converted to maximize the likelihood function:$$logP(\textbf{H}| \mathcal{S})\propto logP(\mathcal{S}| \textbf{H})P(\textbf{H})=\sum_{s_{i,j}\in \mathcal{S}}\omega_{i,j}logP(s_{i,j}|h_{i},h_{j})+\sum_{i=1}^{N}logP(h_{i})$$ where $P(\mathcal{S}|\textbf{H})=\prod_{s_{i,j}\in \mathcal{S}}[P(s_{i,j}|h_{i},h_{j})]^{\omega_{ij}}$,$P(s_{i,j}|h_{i},h_{j})=\sigma(d(h_{i},h_{j}))^{s_{ij}}(1-\sigma(d(h_{j},h_{j})))^{1-s_{ij}}$,$\sigma$ is probability function.

Most deep hashing methods adopt sigmoid function $ \sigma=\frac{1}{1+e^{-\alpha x}}$ as probability function. DCH proposed to use a novel probability function based on Cauchy distribution so as to concentrate relevant images to be within a small Hamming ball and make efficient Hamming space retrieval possible.$$\sigma(d(h_{i},h_{j}))=\frac{\gamma}{\gamma+d(h_{i},h_{j})}$$

\section{Comments}
The whole framework of deep supervised hashing methods for image retrieval can be categorized into three main streams. First, address the ill-posed gradient problems caused by the sign function. Second, design various of loss functions to improve the hash code learning and to make full use of the supervised feature information. Third, design different structures of networks to reduce the redundancy and to recover loss of features during CNN learning.

Comments:

1. It seems that most deep supervised hashing methods focus too much on the semantic features to generate hash codes but ignore texture or spatial features. We have to admit that to achieve better retrieval performance, spatial details should be highly taken into account. For example, if we have a query point which is a picture of a people wearing Nike shoes, we probably retrieve results of people wearing shoes with various of brands. As we know, throughout the CNN architecture, the early convolutional layer tends to retain spatial details asymptotically while the last convolutional layer can capture more semantic and less spatial details. If we extensively exploit features from different convolutional layers, we will probably achieve better performance.

2. Multimedia data on the Internet exists as a range of different types and are represented by different modalities. Since the features that CNNs produce are sometimes too general, cross-modal retrieval should be more suitable for practical application.[10]

3. When comparing with other deep supervised hashing methods, some authors employ more complex structures on their own model while leaving those simple structures on other models in order to get their own state-of-the-art performances. It's quite unfair. We'd better compare the performance on a common benchmark.

\section{One possible attempt}

\subsection{Notation}
We are given a training set of $N$ images: $I={I_{1},I_{2},..,I_{N}},I_{i}\in R^{d_{1}\times d_{2}\times3}$. $Y={y_{i,j}}$ is the corresponding supervised information, which is the similarity matrix in this case. For image $I_{i}$ and $I_{j}$,$Y_{i,j}=0$ if they are similar, and $Y_{i,j}=1$ otherwise. We denote the binary network output of $I_{i}$ by $b_{i}$,where $b_{i}\in \{ \pm 1\}^{k}$, $k$ is the length of hashing codes.

Our goal is to give a map from $I$ to $k$-bit binary codes:  $F: I_{i} \rightarrow b_{i} $ , such that the distances of hashing codes from similar pairs are small, while large for dissimilar pairs.

\subsection{Motivation}
The purpose of supervised hashing is to preserve the small Hamming distances between hashing codes from similar images. As a result, we should minimize the following loss function:
$$L(Y,B)=\sum_{i,j} \frac{1}{2}(1-y_{i,j})H(b_{i},b{j})+\frac{1}{2}max(m-H(b_{1},b_{2}),0)$$
$$s.t. \quad  b_{i}\in \{ \pm 1 \}^{k},\quad 1\leqslant i\leqslant N  \quad $$

However, we cannot directly optimize (1) through back propagation due to the discrete constraints. To fulfill the discrete constrains, lots of supervised methods leverage a common relaxation scheme, which is to replace $b_{i}$ by $sigmoid(b_{i})$ or $tanh (b_{i})$ and get the final hashing codes by $signum$ function. Nevertheless, such nonlinear functions will definitely slow down the learning process[222] and the $signum$ function may lead to suboptimal results. To tackle this problem, a good idea is to transfer the discrete constraints into the regularizer terms. Specifically, a loss function is designed as:
\begin{equation}
\begin{split}
L(Y,B)=& \sum_{i,j}\frac{1}{2}(1-y_{i,j})\left \| b_{i}-b{j}\right \| _{2}^2+\frac{1}{2}y_{i,j}max(m-\left \| b_{i}-b_{j}\right \| _{2}^2,0)\\
&+\alpha (\left \| \left| b_{i}\right| -\mathbf{1}\right \| _{1}+\left \| \left| b_{j}\right| -\mathbf{1}\right \| _{1}) 
\end{split}
\end{equation}
$$s.t. \quad b_{i}\in R^{k}$$

Herein, L2-norm is used to measure the distances between pairs instead of Hamming distance because the subgradient of Hamming distance will treat pairs with different distances equally. Although this model achieves relatively good performance, the disadvantage is obvious. We should notice that the absolute value of subgradient of the regularizer always stays at $\alpha \cdot \mathbf{1}$. During training, when the first term is becoming relatively small compared to the regularizer, the subgradient causes the loss function create jump movements. As a result, the loss function will be basically caught at a value and fails to be optimized further.

\subsection{Model description}
\quad Given that we have gave up utilizing nonlinear relaxation in loss function and regularizer with subgradient like (1), how can we express the discrete restriction in the loss function?

We propose the concept: the "shadow" $u_{i}$ of the network output $b{i}$. During training, we use the "shadow" to instruct the learning of $b_{i}$. In this process, "shadow" stays fixed until this epoch of training ends. Then in the next step, the "shadow" is determined according to the output of the neural network. In other word, $b_{i}$ and its shadow $u_{i}$ are guiding each other in the whole process. Specifically, our loss function is defined as:
\begin{equation}
\begin{split}
L(Y,B)=& \sum_{i,j}\frac{1}{2}(1-y_{i,j})\left \| b_{i}-b{j}\right \| _{2}^2+\frac{1}{2}y_{i,j}max(m-\left \| b_{i}-b_{j}\right \| _{2}^2,0)\\
&+\frac{\alpha}{2}\sum_{i}\left \| b_{i}-u_{i}\right \|_{2}^2+\frac{\beta}{2}\sum_{i}(b_{i}b_{i}^{T}-k)^2 
\end{split}
\end{equation}
$$s.t. \quad b_{i}\in R^{k},\quad u_{i}\in\{ \pm 1\}^{k} $$

During optimization, we sequentially update the parameters of the deep neural network and binary matrix $U$ in the following two steps:
First, update weights with U fixed: we use back propagation to update the neural network parameter. In each iteration, we randomly sample a minibatch of images and use BP algorithm to update the weights. The gradient of the loss with respect to $b_{i}$ follows:
$$\frac{\partial L}{\partial b_{i}}=\sum_{j}(1-y_{i,j})(b_{i}-b_{j})+y_{i,j}(b_{i}-b_{j})+\alpha (b_{i}-u_{i})+\beta (b_{i}b_{i}^{T}-k)b_{i} \quad if \left \| b_{i}-u_{i}\right \|_{2}^{2} \le m$$ 
$$\frac{\partial L}{\partial b_{i}}=\sum_{j}(1-y_{i,j})(b_{i}-b_{j})+\alpha (b_{i}-u_{i})+\beta (b_{i}b_{i}^{T}-k)b_{i} \quad if \left \| b_{i}-u_{i}\right \|_{2}^{2} \ge m \quad(3)$$ 

In the second step, we update U according to instant B. When B is fixed ,(2) becomes: 
$$ min \tilde{L}=\left \| U-B\right \|_{F}^{2}=\mathop{tr}\limits_{U} (U-B)(U-B)^{T}=\mathop{tr}\limits_{U} (U^{T}B) \quad s.t. \quad u_{i}\in \{ \pm 1\}^{k}$$
So we have,$U=sign(B) \quad (4)$.

We iteratively solve the above two steps until the loss function reaches the optimal solution. The algorithm can be  summarized as Algorithm1.

\begin{algorithm}[htb]
 \caption{Framework of Shadow Recurrent Hashing}
 \begin{algorithmic}[1]
  \REQUIRE
  $N$ datapoints $I$;
  supervised similarity matrix$S$;
  hashing code length $k$;
  parameters $\alpha , \beta$;
  number of iteration $T$;
  batch size $M$
  \ENSURE
  hashing codes B
  \STATE Initialize weights,$U$
  \FORALL{$i=1:T$}
   \FORALL{$n=1:\frac{N}{m}$}
   	\STATE Randomly sample a mini-batch from $I$
   	\STATE Calculate $b_{i}$ by forward propagation
	\STATE Calculate the gradient according to (3)
	\STATE Update weights by BP
   \ENDFOR
   \STATE Update U according to (4)
  \ENDFOR
 \end{algorithmic}
\end{algorithm}

\section{Experiment}
\quad  \textbf{Network parameters}:

 I tested the SRH method on CIFAR10 with GPU Geforce 920MX. The SRH method is implemented with Pytorch framework. The network structure consists of three convolution-pooling layers and two fully connected layers. The convolution layers use $32,32,64 5\times 5$ filers with stride 1, and the pooling layers use $3\times3$ windows with stride 2. The first fully connected layer has 500 nodes.The second one has $k$ nodes, where $k$ denotes the length of the hash codes. All activation function are Relu. m is set to be $2\times k$.

I initialize the weight layers with the "Xavier" initialization. During the whole training process, the mini-batch size is set to 160, momentum is set to 0.9, weight decay rate is set to 0.004. The initial learning rate is $10^{-3}$. The margin threshold $m$ is set to $m=2k$ to make sure that more than $\frac{k}{2}$ bits in hash codes are different for dissimilar pairs.

\textbf{Training methodology}:

If we directly train the network with $n$ query points and $\frac{n(n-1)}{2}$ pairs, storing the image pairs could be very consuming and wasted and a common CPU cannot hold such huge batch size. So to make better use of the computational resources and storage place, I propose to generate hash codes of image pairs by exploiting all the image pairs in each mini-batch. As for the image pairs across different batches, we cover the tradeoff by randomly selecting the mini-batches in each iteration and run the whole algorithm plenty of times. As a result, we avoid the inconvenience of dealing with all pairs as a whole, thus my method can be scalable to large-scale datasets.

\textbf{Dataset~CIFAR-10}

This dataset is made up of 60000 images with $32\times 32$ pixels each, which belong to 10 different categories. These images are directly vectorized and used as the input of the CNN architecture. Moreover, we can get the semantic information matrix $S$ from the direct category indicator. Images from the same category are considered to be semantically similar, and vice versa.

\begin{table}[!htbp]
\centering
\begin{tabular} {c|c}
\hline
Models(with $epoch=150,\beta =0.01$) & CIFAR10\\
\hline
$\alpha / \beta =0.1$ & 0.5531\\
$\alpha / \beta =1$ & 0.6166\\
$\alpha / \beta =10$ & 0.5666\\
$\alpha / \beta =100$ & 0.1935\\
\hline
\end{tabular}
\caption{mAP of models, all hashing codes are 12 bits.}
\end{table}

\section{Conclusion}

In this paper, I evaluate several deep supervised hashing methods according to the timeline. Different methods give solution to different existing problems in deep hashing. Some of them has achieved the state-of-the-art performance. Moreover, to break through the bottleneck of existing deep hash method, I propose a new idea "shadow" to accelerate the optimization process. SRH elaborately convert the discrete constraints in loss function to regularizer and design a new type of regularizer to make sure the regularizer term stay calm during training process. Experiments on CIFAR-10 shows the satisfying performance of the proposed method.

\end{document}